\documentclass[a4paper]{jpconf}
\usepackage{graphicx}
\usepackage{subfig}
\usepackage{array}
\usepackage{subfig}
\usepackage{graphicx}
\usepackage{multirow}
\usepackage{multicol}
\usepackage{algorithm}
\usepackage{amsmath}
\usepackage{algorithm}
\usepackage{comment}
\usepackage{enumitem}
\usepackage {algorithm, algorithmic}
\usepackage{hyperref}
\hypersetup{colorlinks = true, allcolors = blue}

\begin{document}
\title{Deep Learning Based Model for Breast Cancer Subtype Classification}

\author{Sheetal Rajpal}
\address{Department of Computer Science, University of Delhi, India}
\ead{sheetal.rajpal.09@gmail.com}

\author{Virendra Kumar}
\address{Department of Nuclear Magnetic Resonance Imaging, All India Institute of Medical Sciences, New Delhi, India}
\ead{virendrakumar@aiims.edu}

\author{Manoj Agarwal}
\address{Department of Computer Science, Hans Raj College, University of Delhi, Delhi, India}
\ead{manoj.agarwal@hrc.du.ac.in}

\author{Naveen Kumar}
\address{Department of Computer Science, University of Delhi, India}
\ead{nkumar@cs.du.ac.in}

\begin{abstract}
Breast cancer has long been a prominent cause of mortality among women. Diagnosis, therapy, and prognosis are now possible, thanks to the availability of RNA sequencing tools capable of recording gene expression data. Molecular subtyping being closely related to devising clinical strategy and prognosis, this paper focuses on the use of gene expression data for the classification of breast cancer into four subtypes, namely, Basal, Her2, LumA, and LumB. In stage 1, we suggested a deep learning-based model that uses an autoencoder to reduce dimensionality. The size of the feature set is reduced from 20,530 gene expression values to 500 by using an autoencoder. This encoded representation is passed to the deep neural network of the second stage for the classification of patients into four molecular subtypes of breast cancer. By deploying the combined network of stages 1 and 2, we have been able to attain a mean 10-fold test accuracy of 0.907 on the TCGA breast cancer dataset. The proposed framework is fairly robust throughout 10 different runs, as shown by the boxplot for classification accuracy. Compared to related work reported in the literature, we have achieved a competitive outcome. In conclusion, the proposed two-stage deep learning-based model is able to accurately classify four breast cancer subtypes, highlighting the autoencoder's capacity to deduce the compact representation and the neural network classifier's ability to correctly label breast cancer patients.
\end{abstract}

\section{Introduction}
\par Breast cancer is a multifaceted condition characterized by the uncontrolled proliferation of abnormal cells. It is a prominent cause of death amongst women, and has surpassed even lung cancer in terms of the maximum number of cases diagnosed in the year 2020 \cite{sung2021global}. It is a genetic disease caused by accumulation of alterations at different levels such as genome, epigenome, and transcriptome. Thus, it becomes important to examine the disease at these levels which was previously impossible. The development of next-generation sequencing technologies has enabled us to take steps towards this direction \cite{reis2009next,sotiriou2009gene}.  Since transcriptome data encoding information in the form of gene expression is capable of capturing maximum variations, a lot of research groups are experimenting with this data. However, high dimensionality still pose challenge to the community.   

\par Breast cancer being such a diverse disease, it has been classified into varoius subtypes \cite{parker2009supervised}. One of the classification category is TNM staging. The T Category distinguishes between the tumors based on their size, the N Category distinguishes them based on invasion of nearby lymph nodes, and the M category determines the metastatic spread of breast cancer to distant organs. Thus, based on cumulative assessment, the cancer stage is labeled as 0, I, II, III, or IV (most severe). Breast cancer may also be classified as per histological grading indicating level of divergence of tumor cells from normal cells. Accordingly, grades 1, 2, and 3 (highest severity) are assigned, based on the severity of the disease. Amongst all categories, molecular subtyping has proven to be more favorable in terms of clinical and prognostic outcomes. This categorization defines four subtypes, namely, Basal, Her2, Luminal A and Luminal B. These subtypes have been defined through PAM50 as well as three Immunohistochemistry (IHC) markers, namely, Estrogen Receptor (ER), Progesterone Receptor (PR), and human epidermal growth factor receptor 2 (Her2). Since, PAM50 defined subtypes have also gained popularity in literature proven to be better associated with outcomes, and thus, gained wider attention \cite{parker2009supervised,cheang2009ki67,perou2000molecular,sorlie2003repeated,kim2019discordance,eccles2013critical}.

\par Choosing a successful breast cancer treatment approach is critically dependent on finding the proper subtype \cite{prat2015clinical,dai2015breast}. Numerous classification approaches such as logistic regression, gradient boosting, support vector machine, bayes classifier, and random forest have been adopted in literature in this regard \cite{wu2017pathways,gao2019deepcc,list2014classification}. With the introduction of deep learning techniques, it has transformed the way how data is processed. These techniques have established themselves in medical domain, more specifically cancer, finding several applications \cite{yan2016comprehensive,kim2016deep,spanhol2016breast,xu2017large,karabulut2017discriminative,ibrahim2014multi,danaee2017deep,fakoor2013using,chen2016gene,singh2016deepchrome}. The goal of this work is to present a deep learning methodology for breast cancer subtype classification.

\par A lot of research is centered on employing IHC marker identified subtypes for breast cancer classification \cite{wu2017pathways,graudenzi2017pathway,sherafatian2018tree,tao2019classifying}. For example, Wu et al. \cite{wu2017pathways} performed large-scale co-expression analysis (using variance analysis) on TCGA BRCA data and identified a set of 136 relevant genes that expressed differentially amongst the four breast cancer subtypes. The functional enrichment and Genomes analysis led to the identification of six functional pathways regulated by the selected genes. Using a support vector machine, they successfully classified the breast cancer subtypes based on these functional pathways. A similar pathway-based analysis was also done by Graudenzi et al. \cite{graudenzi2017pathway} who have used pathway enrichment analysis for feature selection using the TCGA dataset. Subsequently, these differentially expressed genes were tested on different dataset-GEO for the breast cancer subtype classification using Support Vector Machine. Several works in this area have also considered different omics data apart from RNA Sequence data. Islakogl et al. \cite{oztemur2018meta} used miRNA profiles for further categorizing  breast cancer subtypes based on IHC receptors. Using eight independent datasets, they demonstrated the effectiveness of their ranking based approach to determine meta-miRNA corresponding to each breast cancer subtype. However, because of the potential advantage of PAM50 over IHC determined subtypes, we focussed on comparison with PAM50 based breast cancer classification into four subtypes. As a result, we were able to identify single work by Zhang et al. \cite{zhang2018lncrna}. Zhang et al. studied the contribution of long non-coding RNAs (lncRNAs) in classification of breast cancer into different subtypes using TCGA RNAseq data.  They discovered a collection of lncRNAs that they utilised along with chosen coding genes and PAM50 genes, and achieved a classification accuracy of 0.956.  Using a comparatively smaller set, authors have been attained 0.876 and 0.885 for the four-class classification problem using SVM classifier.

\par The suggested approach makes use of deep learning based methodology to stratify breast cancer patients into four subtypes \cite{rajpal2021triphasic}. We presented a model based on deep learning. It employs an autoencoder for dimensionality reduction in stage 1. Autoencoder employed reduces the dimensions from 20,530 gene expression values to a compact encoded representation of size 500. In stage 2, this reduced representation is passed to the deep neural network for classification of patients into four breast cancer subtypes. On deploying the combined network of stage 1 and 2, we have been able to obtain mean 10-fold test accuracy of 0.907. In comparison to the literature, our framework is a pure deep learning approach in which the classifier neural network is developed. Further, we have dealt with the imbalanced nature of the data using Synthetic Minority Over-sampling Technique(SMOTE).

\par Rest of the manuscript is presented as follows: the proposed approach is discussed in the second section. The third section discusses information about the experiments, outcomes, and gene set analysis. The final section concludes the paper and gives the direction for future work. 

\section{Dataset and Methodology}

The section begins with the dataset description followed by  the deep neural network classifier built for categorization of breast cancer into four subtypes.

\subsection{Dataset}

For our research, breast cancer dataset hosted by The Cancer Genome Atlas (TCGA) repository	 is used.  We utilized gene expression data (transcriptome data) of 20,530 genes for 1218 patients and the PAM50 defined subtypes. Since the information about PAM50 defined labels was available for only 837 samples, we  removed remaining samples from further consideration. As a result, we have finally used 147 samples of Basal subtype, 67  of Her2, 434 of Luminal A subtype and 194 samples of Luminal B subtype.

\begin{figure}[!htbp]
\centering
	\subfloat[Autoencoder comprising encoder and decoder network]{
	\includegraphics[width=1.4in,height=6in]{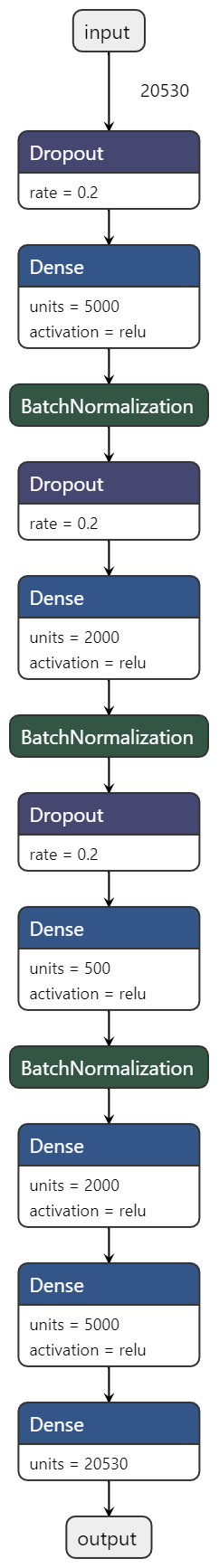}
	\label{fig:03a}
	}
	\hspace{2cm}
	\subfloat[Neural Network classifier for classification into four breast cancer subtype]{
	\includegraphics[width=1.4in,height=6in]{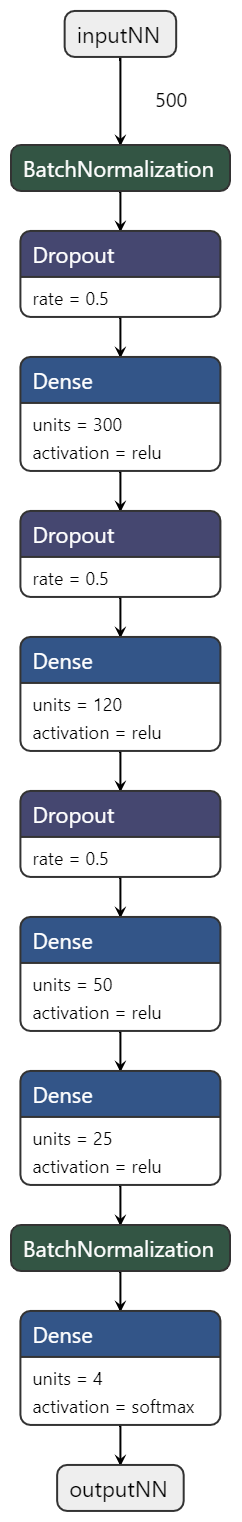}
	\label{fig:03b}
	}

\caption{Architecture  of Deep Neural Networks employed for achieving reduced representation followed by classification using that compressed vector}
\label{fig:03}
\end{figure}

\subsection{Neural Network Classifier Architecture taking 20,530 genes as input}
Deep Neural Network classifier comprises a neural network for reducing the number of dimensions in stage 1, whose input is subsequently passed to a neural network employed for categorization of breast cancer into one of the four breast cancer subtypes in stage 2 (see Figure \ref{fig:03}). In first stage, neural network employed is an autoencoder which comprises first three dense layers with 5000, 2000, and 500 nodes responsible for achieving reduced representation followed by three more dense layers with 2000, 50000, and 20,530 nodes for reconstructing the input values using reduced representation.  While the first three dense layers constitute an encoder, the latter three dense layers are part of the decoder. Output of encoder i.e. feature representation of size 500 is fed as an input in the second neural network employed for classification. It comprises five dense layers with 300, 120, 50, 25, and 4 neurons. Thus, the final combined network deployed for classification comprises encoder and the second neural network.

To facilitate learning by neural network, autoencoder employs batch-normalization which normalizes the outputs (in batches) of dense layers passed to subsequent layers and dropout rate of 0.2 so as to prevent network from overfitting the data. Similarly, the second classifier network also employs batch-normalization with a dropout rate of 0.5.

\section{Experimentation}
The sections discuss the pre-processing that was done on the dataset used for the experiment. Following that, we describe  hyper-parameters tuned for the employed neural networks. Subsequently, classification accuracy of the proposed deep neural network classifier is accessed using the entire set of 20,520 genes.

\subsection{Pre-Processing and Hyperparameters}
For achieving data normalization, z-score normalization is applied to gene expression data Further, the PAM50 defined class labels, namely, Basal, Her2, Luminal A, and Luminal B were mapped to one-hot encoding vectors. Since the distribution of samples across classes is highly imbalanced, we have used Synthetic Minority Over-sampling Technique (SMOTE).

During training, a batch size of 32 is employed. In addition, the Adam optimization technique and a learning rate of 0.0006 are used to aid neural network learning. Furthermore, the initial weights of the network are assigned using the default keras initializer.

\subsection{Classification Performance using 20,530 genes}
\par
\begin{figure}[!htbp]
\centering
	\subfloat[Phase I]{
	\includegraphics[width=0.48\linewidth]{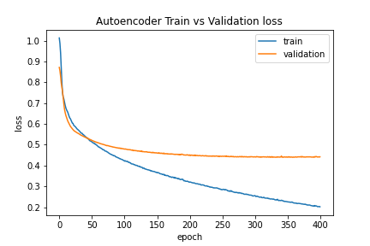}
	\label{fig:04a}
	}
	\hfill
	\subfloat[Phase 2]{
	\includegraphics[width=0.48\linewidth]{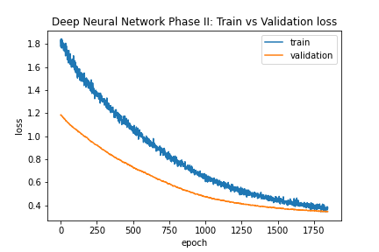}
	\label{fig:04b}
	}
\caption{Change in training loss in phase 1 and phase 2 network
}
\label{fig:03}
\end{figure}

 For the purpose of experimentation, we first pre-processed the data. The dataset is then divided into training (90\% of the samples) and test set (10\% of the samples). We also put aside 10\% of the training set as the hold-out validation set so as to facilitate model development and hyper-parameter setting while avoiding overfitting. The combined network presented in the previous section is leveraged for the classification task. The model is first trained on the training partition, and its classification performance is evaluated on the test partition. The training and validation loss for the first and second phase are depicted in Figure \ref{fig:04a} and \ref{fig:04b} respectively.

\begin{figure}[!t]
\centering
	\includegraphics[width=3.2in,height=5cm]{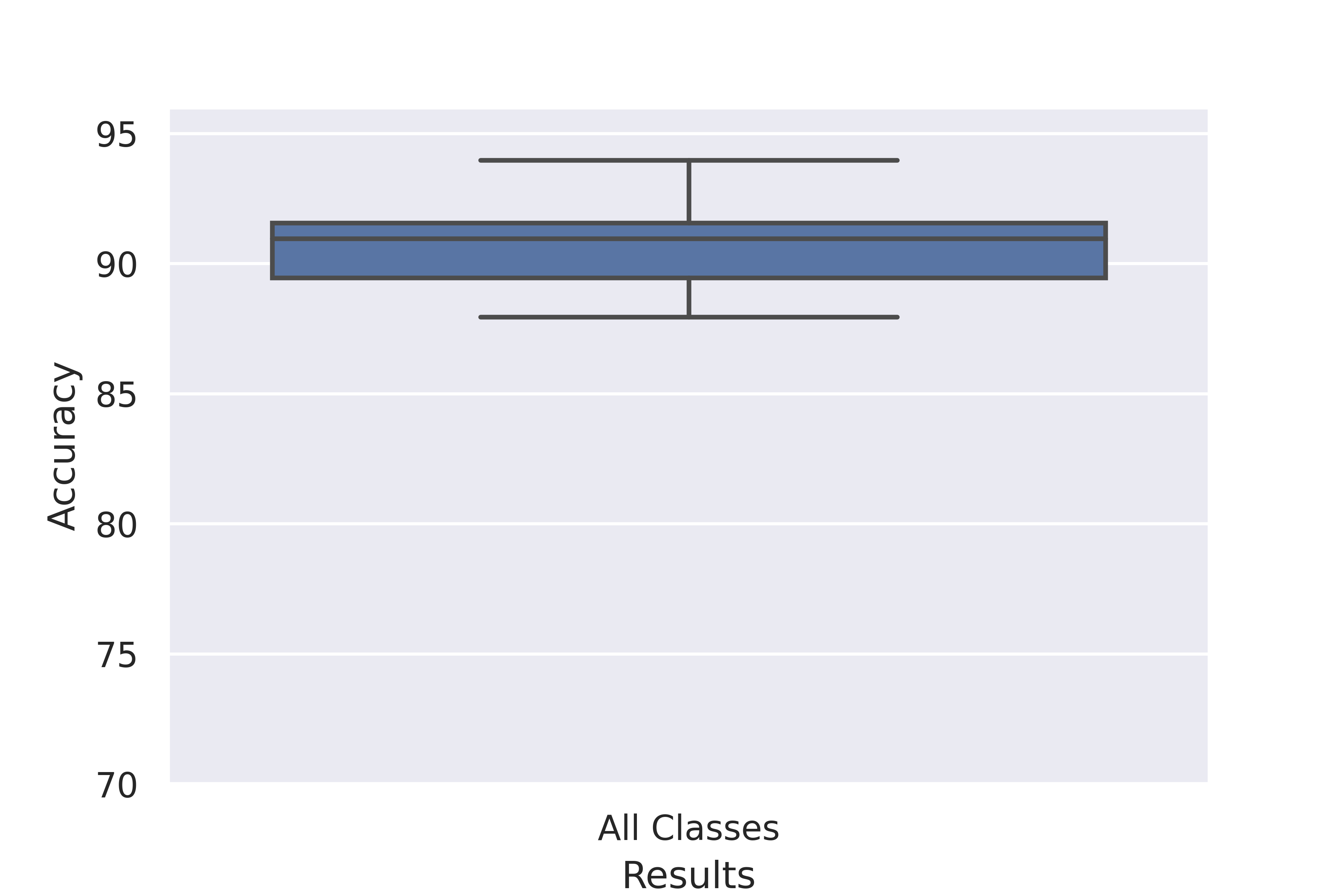}
\caption{BoxPlot shows that the accuracy is quite stable and typically varies only in the range 0.879 to 0.939 for 10-fold cross-validation for the classifier when 20,530 genes are used.}
\label{fig:05}
\end{figure}

\begin{figure}[!t]
\centering
	\includegraphics[width=3.2in,height=5cm]{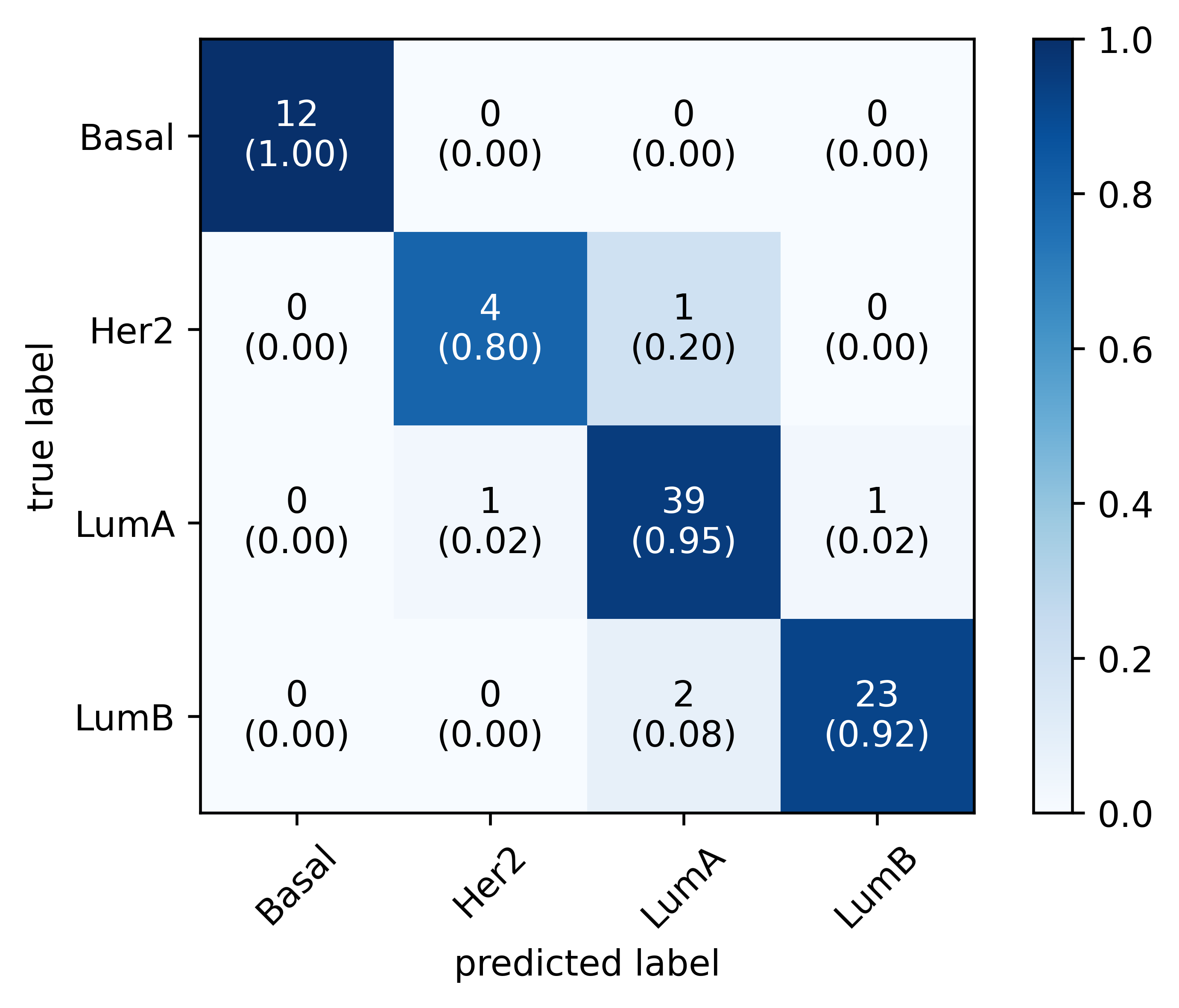}
\caption{Confusion Matrix specifying Classification accuracy across 10 independent runs of the proposed framework (10 fold cross validation accuracy: 90.72\%.}
\label{fig:06}
\end{figure}

We used 10-fold cross-validation to test the stability of the presented classifier model for various training/test split options. We attained a mean 10-fold test accuracy of 0.907 using the proposed approach. Figure \ref{fig:05} shows the variation in accuracy score across 10 distinct folds of the dataset through boxplot visualization. Further, the Figure \ref{fig:06} presents with confusion matrix depicting number of instances correctly classified. The results clearly illustrate stability of the classification performance.


\subsection{Comparison with related work}

\par For comparing the classification performance, we have compared the outcome of the proposed methodology with the work proposed by Zhang et al. \cite{zhang2018lncrna}. They have considered four-class classification problem subtyping breast cancer into Basal, Her2, Luminal A, and Luminal B.  Without explicit inclusion of PAM50 genes, they have been able to obtain 0.885 using 36 genes. However, we have attained a mean 10-fold test accuracy of 0.907 using the proposed approach. Thus, the results obtained are competitive.

\section{Conclusion}

As molecular categorization has proven to be more favourable in terms of clinical and prognostic outcomes, devising an effective treatment therapy requires the subtyping into one of the four breast cancer subtypes, namely, Basal, Her2, Luminal A, and Luminal B. In this direction, we have presented a two-stage deep learning model. In stage 1, an autoencoder is used to reduce dimensionality from 20,530 gene expression values to a compact encoded repression of size 500. The encoded representation is fed into the second stage's deep neural network, which classifies patients into four breast cancer subtypes. We were able to achieve a mean 10-fold test accuracy of 0.907 percent using the combined network of stages 1 and 2. This clearly illustrates the capability of the proposed two-stage deep learning-based model to distinguish between four breast cancer subtypes using gene expression data. The boxplot shows that the suggested framework's classification accuracy is quite consistent throughout 10 separate runs. Further, we also achieved competitive results when compared to related work.  In summary, the proposed two-stage deep learning-based model is capable to identify four breast cancer subtypes emphasizing the autoencoder's capacity to derive the compact representation and the neural network classifier's ability to accurately label breast cancer patients. As part of future work, we aim to apply the proposed model for classifying different types of cancers. Further, we intend to incorporate genome and epigenome data also along with transcriptome data for studying this heterogeneity.

\section{References}

\bibliographystyle{iopart-num}
\bibliography{document}

\end{document}